\documentclass[10pt,twocolumn,letterpaper]{article}

\usepackage{iccv}
\usepackage{times}
\usepackage{epsfig}
\usepackage{graphicx}
\usepackage{amsmath}
\usepackage{amssymb}
\usepackage{booktabs}
\usepackage{color}
\usepackage{subfigure}
\usepackage{multirow}
\usepackage{makecell}
\usepackage{bbding}
\usepackage{xcolor}
\usepackage{bm}

\usepackage[pagebackref=true,breaklinks=true,letterpaper=true,colorlinks,bookmarks=false]{hyperref}
\iccvfinalcopy 

\def\iccvPaperID{8867} 

\ificcvfinal\fi

\begin{document}

\title{VIMI: Vehicle-Infrastructure Multi-view Intermediate Fusion \\for Camera-based 3D Object Detection}

\author{Zhe Wang$^{1}$, Siqi Fan$^{1}$, Xiaoliang Huo$^{1,2}$, Tongda Xu$^{1}$, Yan Wang$^{1}$\thanks{Corresponding authors.} , Jingjing Liu$^{1}$, Yilun Chen$^{1}$, Ya-Qin Zhang$^{1}$\footnotemark[1]\\
$^{1}$ Institute for AI Industry Research (AIR), Tsinghua University, Beijing, China \\
$^{2}$ Beihang University, Beijing, China \\
{\tt\small \{wangzhe, fansiqi, wangyan\}@air.tsinghua.edu.cn}
}

\maketitle
\ificcvfinal\thispagestyle{empty}\fi

\renewcommand{\thefootnote}{\fnsymbol{footnote}}
\footnotetext[2]{Code will be made publicly available at the \href{https://github.com/Bosszhe/VIMI}{link}.}
\begin{abstract}

In autonomous driving, Vehicle-Infrastructure Cooperative 3D Object Detection (VIC3D) makes use of multi-view cameras from both vehicles and traffic infrastructure, providing a global vantage point with rich semantic context of road conditions beyond a single vehicle viewpoint. 
Two major challenges prevail in VIC3D: $1)$ inherent calibration noise when fusing multi-view images, caused by time asynchrony across cameras;
$2)$ information loss when projecting 2D features into 3D space.
To address these issues, We propose a novel 3D object detection framework, Vehicles-Infrastructure Multi-view Intermediate fusion (VIMI).
First, to fully exploit the holistic perspectives from both vehicles and infrastructure, we propose a Multi-scale Cross Attention (MCA) module that fuses infrastructure and vehicle features on selective multi-scales to correct the calibration noise introduced by camera asynchrony. Then, we design a  Camera-aware Channel Masking (CCM) module that uses camera parameters as priors to augment the fused features. We further introduce a Feature Compression (FC) module with channel and spatial compression blocks to reduce the size of transmitted features for enhanced efficiency.
Experiments show that VIMI achieves 15.61\% overall $AP_{\text{3D}}$ and 21.44\% $AP_{\text{BEV}}$ on the new VIC3D dataset, DAIR-V2X-C, significantly outperforming state-of-the-art early fusion and late fusion methods with comparable transmission cost.

\end{abstract}
\section{Introduction}

3D object detection is one of the most important environmental perception tasks for autonomous driving (AD). 
Subject to sensor limitations, autonomous vehicles lack a global perception capability for monitoring holistic road conditions and accurately detecting surrounding objects, which bears great safety risks. Vehicle-to-everything (V2X) aims to build a communication system between vehicles and other devices in a complex traffic environment. V2X can further enlarge the perception range of a single vehicle and enables detection for blind areas.



\begin{figure}[t]
	\centering  
	\includegraphics[width=\linewidth]{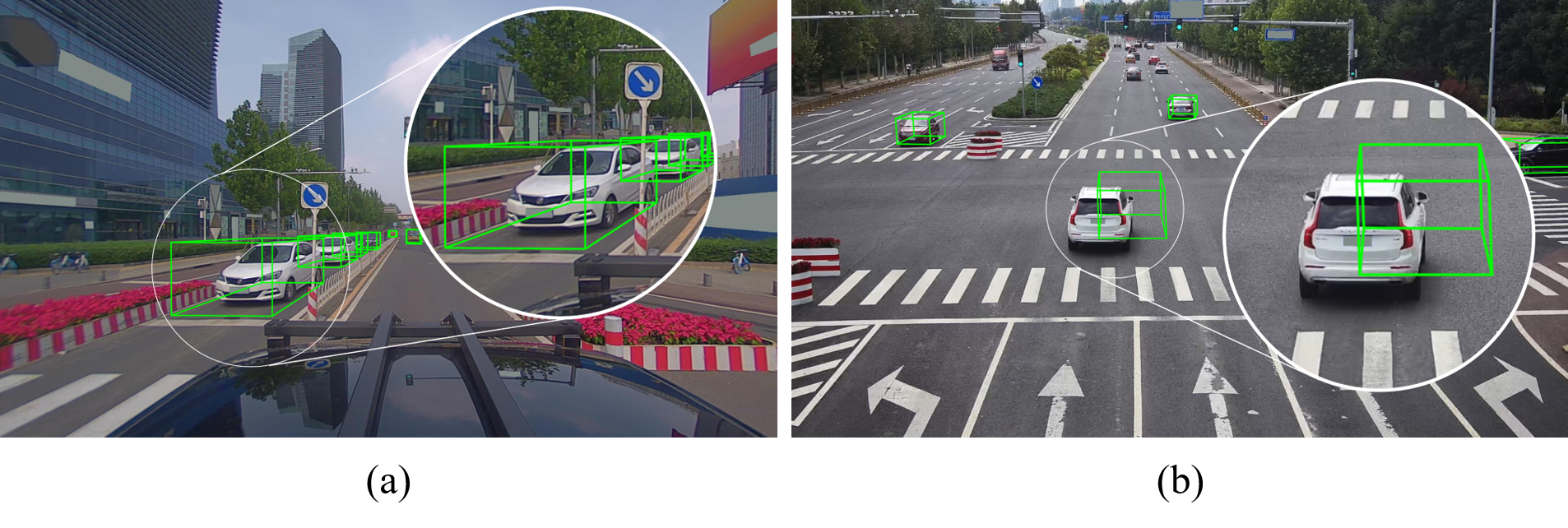} 
	\caption{Labels (3D bounding boxes) projected from 3D space to vehicle (a)  and infrastructure (b) image planes often suffer from misalignment between the ground truth and the projection position in 2D images (as illustrated by the misaligned green bounding boxes), because calibration noise inherently exists in the joint-labeling of different views in VIC3D datasets.}  
	\label{fig:calib_noise}   
\end{figure}

Existing public V2X datasets are mostly simulated, such as OPV2V~\cite{xu2022opv2v}, V2X-Sim~\cite{li2022v2xsim} and V2XSet~\cite{xu2022v2xvit}. Most existing research on V2X has focused on LiDAR-assisted methods, including \textit{early fusion} (EF) of raw signals~\cite{yu2022dairv2x,hu2022where2comm,chen2022co3}, \textit{intermediate fusion} (IF) of features~\cite{mehr2019disconet,xu2022opv2v,wang2020v2vnet}, and \textit{late fusion} (LF) of prediction outputs~\cite{yu2022dairv2x,chen2022model-agnostic}. Most recent research~\cite{yu2022dairv2x,chen2022model-agnostic} adopts a late-fusion method based on 3D predictions of each monocular detector (e.g., 3D bounding boxes from the camera and LiDAR). When considering intermediate fusion, prior methods~\cite{wang2020v2vnet,mehr2019disconet,hu2022where2comm,xu2022v2xvit} have mainly focused on additional features extracted from simulated point clouds in Vehicle-to-Vehicle (V2V) scenarios. 
As LiDAR is highly expensive and difficult to deploy in each vehicle in practical applications, an alternative solution is Vehicle-to-Infrastructure (V2I), in which case standard cameras are installed in shared traffic environment providing a holistic view of road conditions. Due to the lack of real V2I infrastructure and publicly available data, few studies have investigated such a vehicle-infrastructure camera fusion problem.
Recently, DAIR-V2X~\cite{yu2022dairv2x} proposed Vehicle-infrastructure cooperative 3D object detection (VIC3D) task and released new benchmarks using point clouds and camera images from real scenarios. These datasets contain real data with roadside cameras complimenting single vehicle viewpoint, which provides a broader perception range that better captures vehicle blind spots. The baseline method in~\cite{yu2022dairv2x} relies on late fusion by combining prediction outputs from each camera. 

In this paper, we propose a novel framework for this new VIC3D task, \textit{Vehicle-Infrastructure Multi-view Intermediate fusion} (VIMI).
We choose intermediate instead of late fusion, as the latter highly relies on accurate values of extrinsic and intrinsic camera parameters. This is not guaranteed in VIC3D task, as there exists an inherent temporal asynchrony caused by transmission delay and calibration noises between the vehicle and infrastructure. As shown in Figure~\ref{fig:calib_noise}, this time asynchrony and calibration error can result in inaccurate relative position detection. 
By focusing on feature-level fusion between vehicle and infrastructure cameras, high-dimensional features extracted from raw data can be compressed and transmitted, which can be used 
to alleviate the negative effect of calibration noises.


Specifically, VIMI includes a Feature Compression (FC) module which compresses 2D features transmitted from the infrastructure to vehicle to alleviate transmission delay. Then, considering the same object can be captured by sensors from both vehicle and infrastructure at different distances, we introduce a Multi-scale Cross Attention (MCA) module to attentively fuse multi-scale features according to feature scale correlations between vehicle and infrastructure. To correct calibration errors born from multiple cameras, features from both infrastructure and vehicle are further enhanced by a  Camera-aware Channel Masking (CCM) module via a learned channel-wise mask following guidance of camera priors (intrinsic and extrinsic parameters). Then, the refined features are transformed into voxel features leveraging calibration parameters and projected into 3D space. Finally splatted into BEV space, the fused feature is fed into detection heads for object detection. For evaluation purposes, we have built a new multi-view camera fusion benchmark on the latest DAIR-V2X dataset. Experiments demonstrate the effectiveness of each VIMI module in reducing calibration error and achieving better prediction accuracy than existing EF and LF methods. 

Our contributions can be summarized as follows:

\begin{itemize}
    \item We propose VIMI, a novel framework for multi-view 3D object detection, the first intermediate-fusion method for camera-based VIC3D task. 

    \item We design MCA and CCM modules to dynamically augment image features for better detection performance, with an additional FC module to reduce transmission costs in VIC3D system.

    \item 
  We achieve state-of-the-art results on DAIR-V2X-C dataset, the latest VIC3D benchmark with real data, where VIMI outperforms existing LF and EF methods with comparable transmission costs.

\end{itemize}

\section{Related Work}

\subsection{V2X Cooperative Perception}

Current research on V2X cooperative perception mainly focuses on simulated datasets, such as OPV2V~\cite{xu2022opv2v}, V2X-Sim~\cite{li2022v2xsim} and V2XSet~\cite{xu2022v2xvit}. Existing intermediate-fusion methods focused on simulated point clouds, such as V2VNet~\cite{wang2020v2vnet} transmitted compressed features to nearby vehicles and generated joint perception/prediction. DiscoNet~\cite{mehr2019disconet} introduced graphs into feature fusion and proposed edge weights to highlight different informative regions during feature propagation. Recent Where2comm~\cite{hu2022where2comm} considered the spatial confidence of features and selected features with high confidence and complementary to others, which effectively saves transmission costs. 
Different from point clouds, images from vehicle and infrastructure have a huge view gap, thus features need to be transformed into unified space for fusion. One direct way for fusing multi-view images is late fusion, such as DAIR-V2X~\cite{yu2022dairv2x}, which proposed a result-level fusion model for cameras with separate detectors~\cite{rukhovich2022imvoxelnet}. Few approaches have focused on IF methods for cameras, especially in real scenarios. 

\begin{figure*}[ht]
	\centering  
	\includegraphics[width=\linewidth]{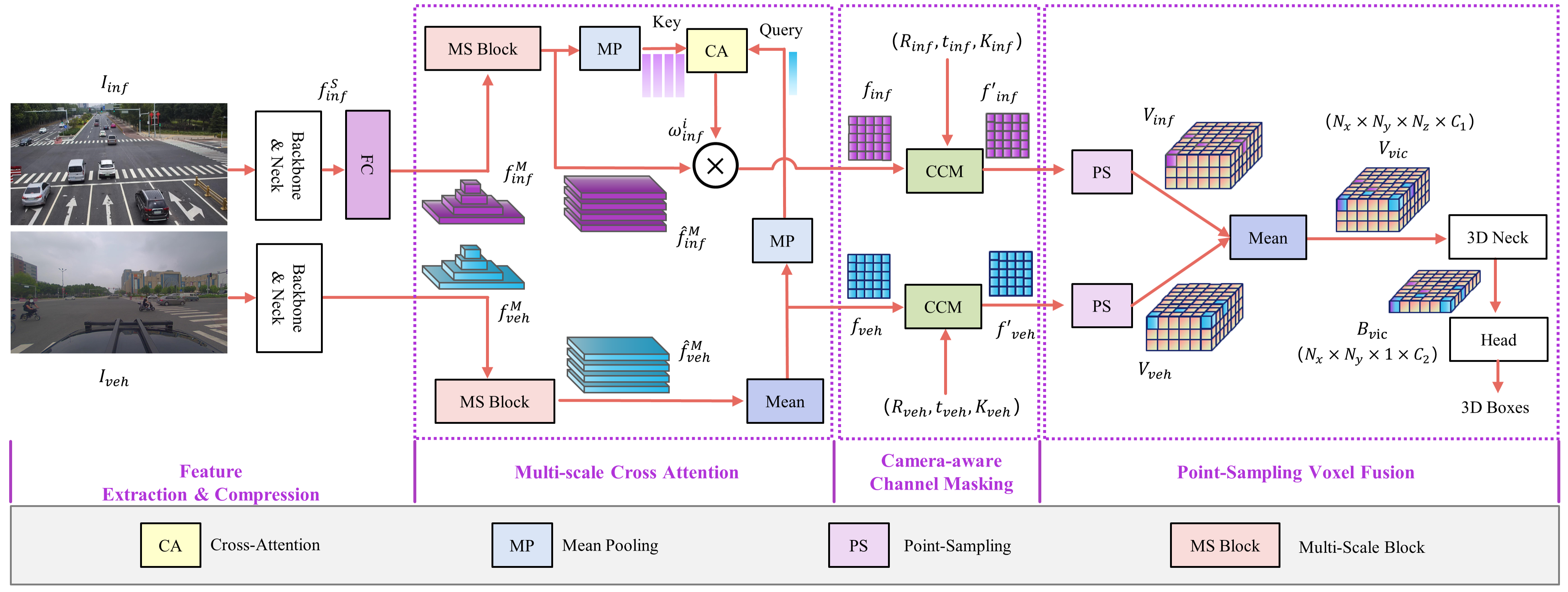} 
	\caption{The general framework of VIMI. Separate image backbone and neck extract multi-scale image feature from vehicle and infrastructure images. FC module compresses source infrastructure feature $f^{S}_{inf}$ and decompresses it to multi-scale ones $f^{M}_{inf}$. MCA module augments features $f^{M}_{veh/inf}$ by seeking the correlation between the two sides, and CCM takes camera parameters $(R,t,K)$ as input to reweight features $f_{veh/inf}$ with channel relationship. Finally, Point-Sampling Voxel Fusion projects image features $f^{\prime}_{veh/inf}$ into 3D space to generate a unified voxel feature $V_{vic}$, which can be applied to 3D neck and head in turn for detection prediction. 
 }  
	\label{fig:framework}   
\end{figure*}

\subsection{Multi-View Camera Fusion}


\textbf{Direct Prediction} methods extract image features with object query~\cite{wang2022detr3d,chen2022futr3d,liu2022petr,liu2022petrv2} or directly on front-view image~\cite{wang2021fcos3d}. DETR3D~\cite{wang2022detr3d} used a sparse set of 3D object queries to sample 2D multi-view image features and predicted 3D bounding boxes with set-to-set loss. PETR~\cite{liu2022petr,liu2022petrv2} transformed image features into 3D position-aware representation by encoding 3D coordinates into position embedding. FCOS3D~\cite{wang2021fcos3d} transformed 3D labels to front-view images and directly predicted 3D information by extending FCOS~\cite{tian2019fcos} to 3D detection.

\textbf{Lift-based} methods project features from image plane to BEV  (bird's eye view) plane through depth estimation. Most methods~\cite{huang2021bevdet,huang2022bevdet4d,xie2022m2bev,zhang2022beverse,reading2021caddn} applied 2D-to-3D transformation following LSS~\cite{philion2020lss}, which predicted a depth distribution for each pixel and lifted image features into frustum features with camera parameters, then splatted all frustums into a rasterized BEV feature. BEVDepth~\cite{li2022bevdepth} claimed the quality of intermediate depth estimation is the key to improving multi-view 3D object detection and added explicit depth supervision with groundtruth depth generated from point clouds. PON~\cite{roddick2020pon} learned the transformation leveraging geometry relationship between image locations and BEV locations in the horizontal direction. 

\textbf{Projection-based} methods generate dense voxel or BEV representation from image features through 3D-to-2D projection~\cite{ma2022bevsurvey}. ImVoxelNet~\cite{rukhovich2022imvoxelnet} aggregated the projected features from several images via a simple element-wise averaging, where spatial information might not be exploited sufficiently.
Transformer-based methods~\cite{li2022bevformer,peng2022bevsegformer} mapped perspective view to BEV with designed BEV queries and leveraged cross- and self-attention to aggregate spatial and temporal information into BEV queries. Since global attention needs huge memory with high time cost, deformable attention was adopted in BEVFormer~\cite{li2022bevformer}.

\section{VIMI Framework}

VIMI aims to fuse vehicle and infrastructure features by utilizing V2X communication. It includes four main modules: Feature Compression (FC), Multi-Scale Cross Attention (MCA), Camera-aware Channel Masking (CCM), and Point-Sampling Voxel Fusion, as illustrated in Figure~\ref{fig:framework}. 

System input is a pair of RGB images from both vehicle and infrastructure cameras. First, features are extracted separately from backbone and 2D neck on each side. Then, the infrastructure feature is sent to FC module, which compresses the feature, transmits it to vehicle side, and decompresses the feature. Multi-scale features are generated from decompression output and sent to MCA module for augmentation. Then, image features are integrated with camera parameters through CCM module. The augmented features are projected into a 3D voxel volume, which aggregates the features via element-wise averaging. Next, the fused voxel feature is transformed into BEV feature via 3D neck. Finally, 2D detection heads composed of several CNN blocks take the BEV feature as input and predict 3D bounding boxes. 

\begin{figure}[t]
	\centering  
	\includegraphics[width=0.8\linewidth]{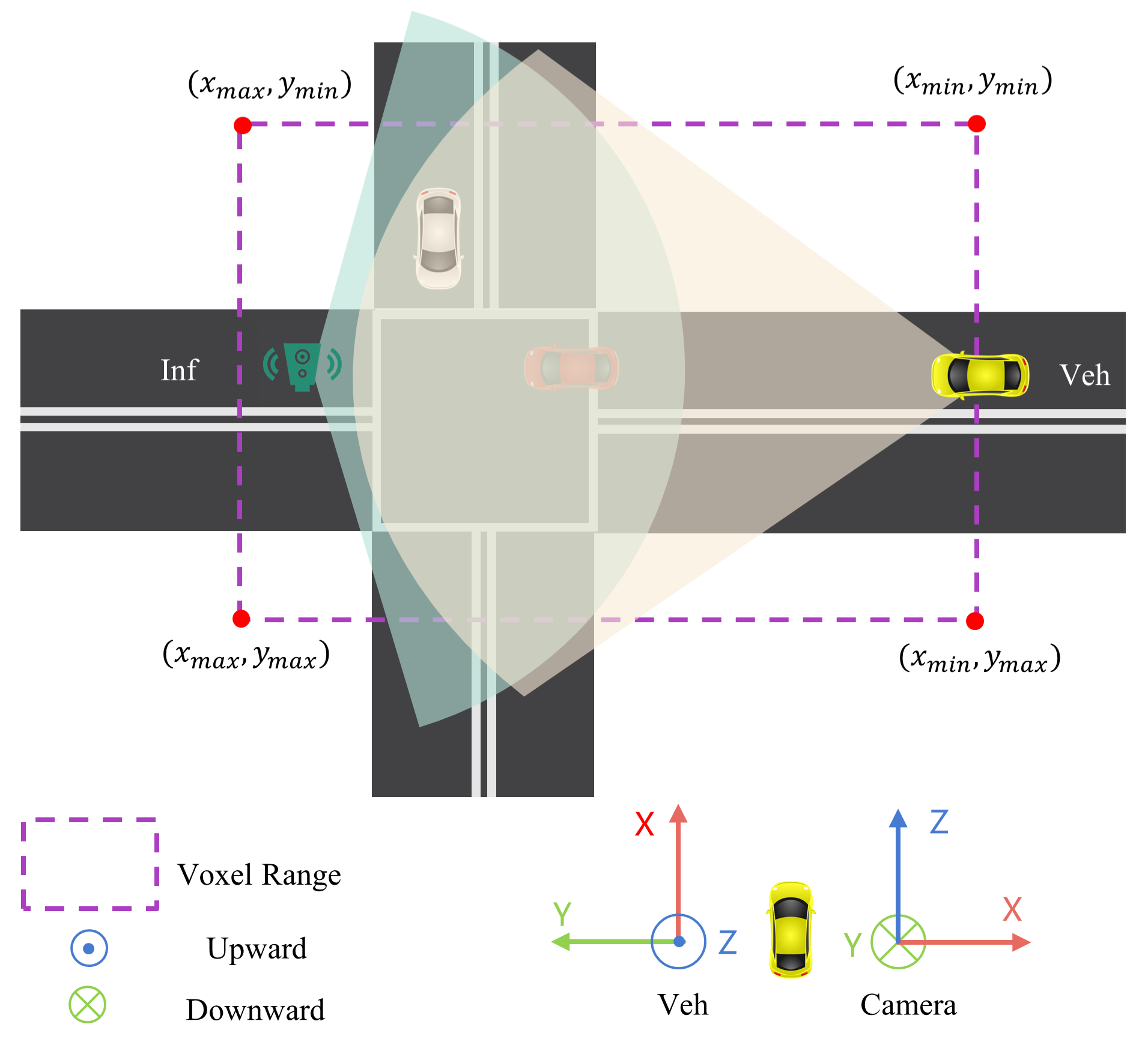} 
	\caption{Illustration of the coordinate system in VIMI from BEV. The vehicle (in yellow) communicates with infrastructure (in green), and the two cameras have different view fields. The vehicle coordinate system takes LiDAR as the origin with $x$- and $y$- axis parallel to the ground and $z$-axis upward vertically. Image features need to be transformed into a voxel range (in purple rectangle), as detailed in section 3.4.
}
	\label{fig:Coord}   
\end{figure}

Prediction results are in ego-vehicle coordinate system, which is shown in Figure~\ref{fig:Coord} and can be parameterized as $(x, y, z, w, h, l, \theta)$, where $(x, y, z)$ are the coordinates of box's center, $w, h, l$ refer to object's width, height and length, and $\theta$ is rotation angle around $z-$axis. 

\subsection{Feature Compression}

\begin{figure}[ht]
	\centering  \includegraphics[width=0.9\linewidth]{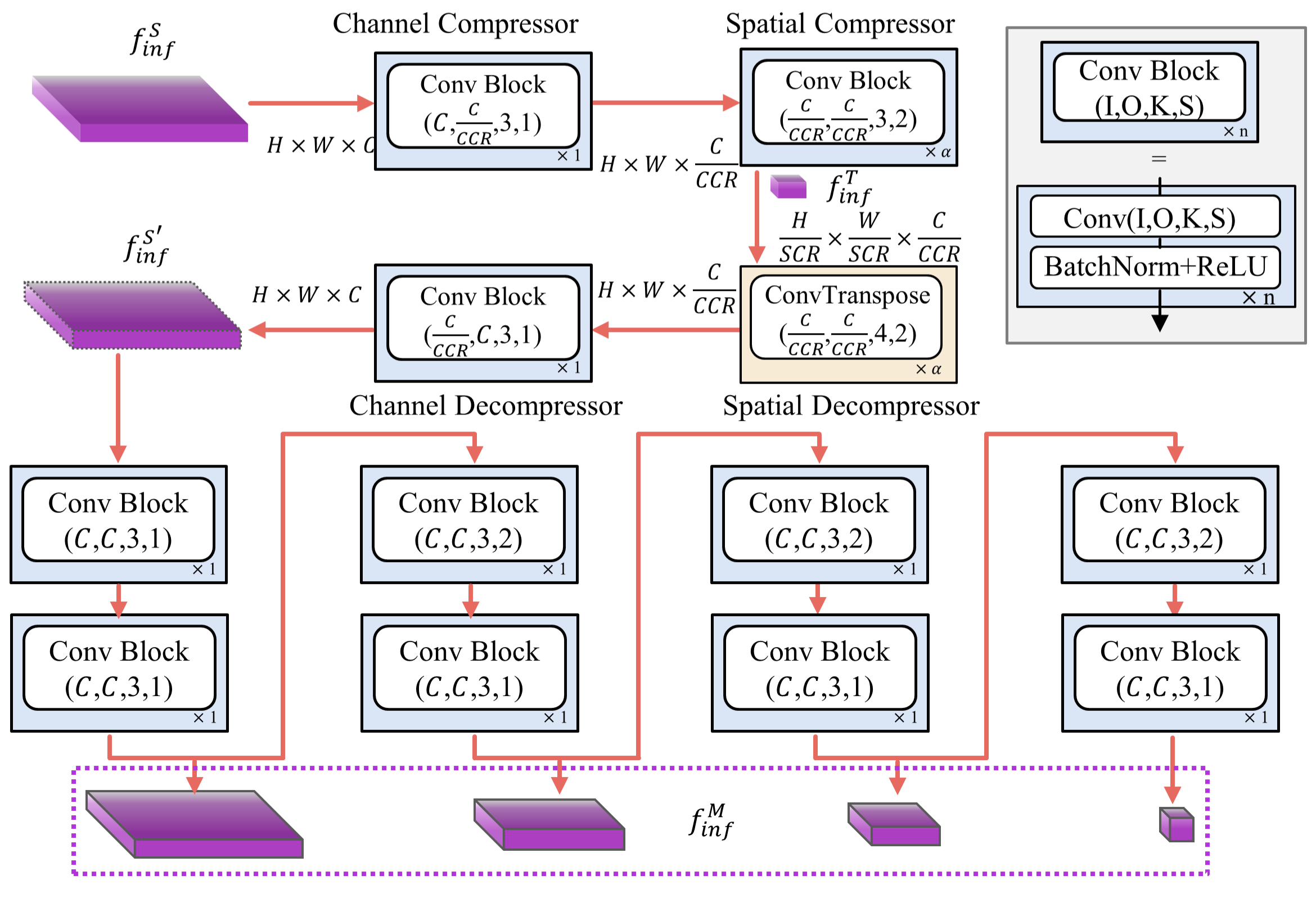} 
	\caption{Illustration of FC module. Feature $f_{inf}^{S}$ is compressed into $f_{inf}^{T}$ through the channel and spatial compressors, which is transmitted to vehicle and is decoded into $f_{inf}^{S\prime}$ through the channel and spatial decompressors. Finally, multi-scale infrastructure features $f_{inf}^{M}$ can be recovered from $f_{inf}^{S\prime}$ with several Conv Blocks with stride 2. 
 }
	\label{fig:FC}   
\end{figure}

The images from vehicle and infrastructure are denoted as $I_{veh}$ and $I_{inf}$, respectively, and the shape of both images are $ \left[ H \times W \times 3 \right]$. We use separate pre-trained backbones and necks on the vehicle and infrastructure respectively to extract multi-scale image features. The output multi-scale features $f^{M}_{s}, s = veh/inf$ can be denoted as:


\begin{equation}
    \begin{split}
        &f^{M}_{s} =\{f^{m}_{s} \in \left[h_m \times w_m \times C \right],\; m=0,1,2,3\} \\
        &h_0=\frac{H}{4},\; w_0=\frac{W}{4},\; h_m=\frac{h_{m-1}}{2},\; w_m=\frac{w_{m-1}}{2}
    \end{split}
\end{equation}

VIMI transmits image features and camera parameters instead of voxel feature after projection because voxel feature is too large to be transmitted efficiently. The Feature Compression (FC) module (shown in Figure~\ref{fig:FC}) compresses the largest infrastructure feature $f^{0}_{inf}$ (noted as $f^{S}_{inf}$), transmits it to vehicle and regenerate multi-scale features$f^{M}_{inf}$ through decompression.


The compression and decompression process in FC module is an Encoder-Decoder with four components: Channel Compressor (CC), Spatial Compressor (SC), Spatial Decompressor (SD), and Channel Decompressor (CD). CC and CD are composed of several convolutional layers. SC comprises several Conv Blocks with stride 2 so that feature scales are reduced to half after each one. SD only replaces convolution with transposed convolution. The Compression Rate (CR) is determined by Channel Compression Rate (CCR) and Spatial Compression Rate (SCR). The number of SC's layers is calculated by $\alpha = \log_{4}{\text{SCR}}$.

\subsection{Multi-scale Cross Attention}

\begin{figure}[htbp]
	\centering  
	\includegraphics[width=0.9\linewidth]{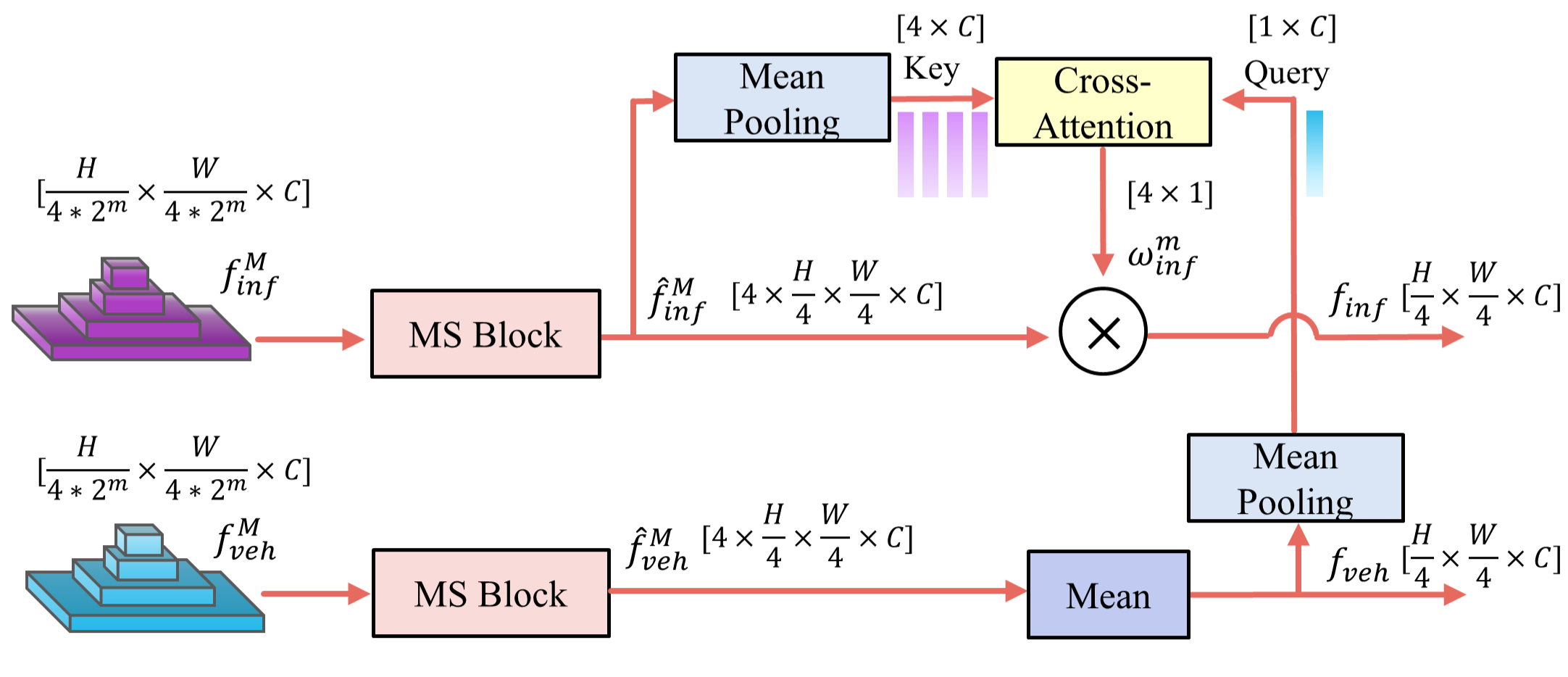} 
	\caption{Schema of MCA module.  In the lower branch, vehicle feature $f_{veh}$ is generated from $f^{M}_{veh}$ through MS Block and Mean. In the upper branch, $f^{M}_{inf}$ is refined into `key' through MS Block and MeanPooling, and queries are generated from $f_{veh}$ through MeanPooling. The output weights $\omega_{inf}^{m}$ of cross-attention are applied to $\hat{f}^{M}_{inf}$ with inner product to form infrastructure feature$f_{inf}$.} 
	\label{fig:MCA}   
\end{figure}

\begin{figure}[ht]
	\centering  
	\includegraphics[width=0.9\linewidth]{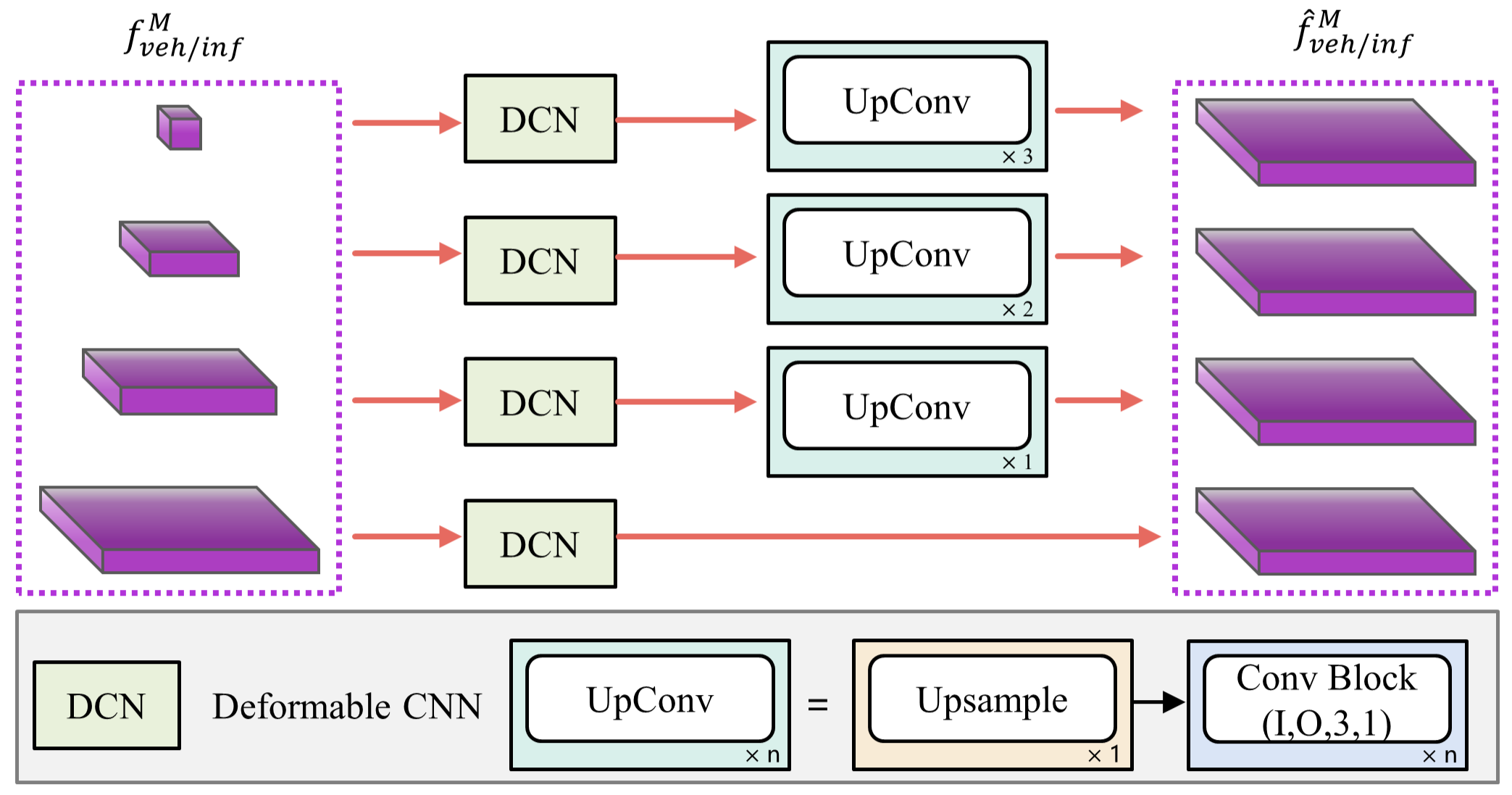} 
	\caption{Details of MS Block. Every pixel-wise feature is integrated with the spatial information of surrounding pixels via DCN, and multi-scale features are scaled to the same size through UpConv blocks.}  
	\label{fig:MSB}   
\end{figure}

MCA module (Figure ~\ref{fig:MCA}) applies cross-attention between multi-scale infrastructure and vehicle features to select useful multi-scale features, and contains a Multi-Scale (MS) Block to alleviate the negative effect of calibration noise. Multi-scale features get spatial information surrounding each pixel and are scaled to the same size through MS Block (Figure~\ref{fig:MSB}).


 MCA applies MeanPooling operation to obtain the representation of different scales of infrastructure features, while vehicle features at different scales are first fused by mean operation and then refined by MeanPooling. To find the correlation between vehicle features and infrastructure features at different scales, cross attention is applied to infrastructure representations as Key and vehicle representation as Query, which generates attention weights $\omega^{m}_{inf}$ for each scale $m$. We calculate inter-product between features $\hat{f}^{M}_{inf}$ and weights $\omega^{m}_{inf}$. The final outputs of MCA are augmented infrastructure image feature $f_{inf}$ and vehicle image feature $f_{veh}$.

\subsection{Camera-aware Channel Masking}


Considering the assumption that the nearer to the camera, the more valuable information can be obtained. And given that camera's extrinsic and intrinsic parameters can instill image features with camera distance information, it is intuitive to take camera parameters as priors to augment image features. 

\begin{figure}[htbp]
	\centering  
	\includegraphics[width=0.9\linewidth]{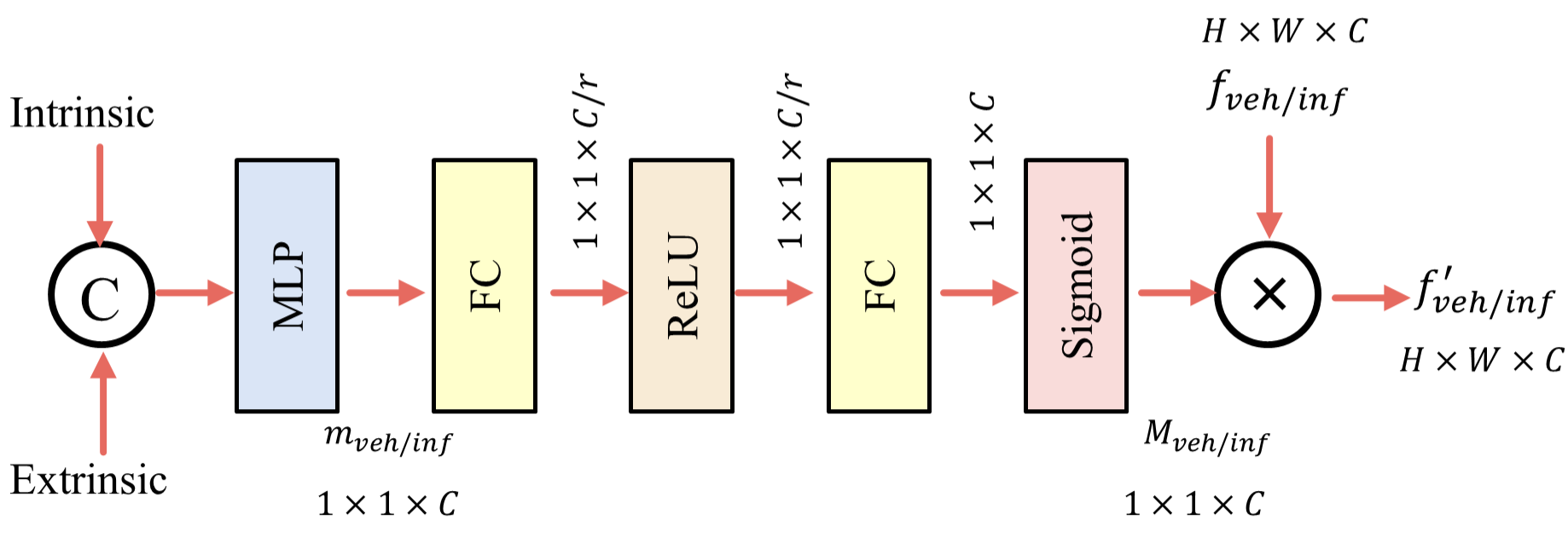} 
	\caption{The schema of CCM module. Camera intrinsic and extrinsic are encoded into the channel mask, then image feature is integrated with it through inner-product operation.}  
	\label{fig:CCM}   
\end{figure}

 Inspired by the decoupled nature of SENet~\cite{hu2018SE} and  LSS~\cite{philion2020lss}, we generate a channel mask to let each feature be aware of camera parameters (Figure~\ref{fig:CCM}). First, camera intrinsic and extrinsic are stretched into one dimension and concatenated together. Then, they are scaled up to the feature’s dimension $C$ using MLP to generate a channel mask $M_{veh/inf}$. Finally, $M_{veh/inf}$ is used to re-weight the image features $f_{veh/inf}$ in channel-wise and obtain results $f^{\prime}_{veh/inf}$. The overall CCM module can be written as:

\begin{equation}
    \begin{split}
        f^{\prime}_{s}  &= M_{s} \odot f_{s} , s= veh, inf \\
        m_{s} &=  \text{MLP} \left(\xi\left(R_s\right) \oplus \xi\left(t_s\right) \oplus \xi\left(K_s\right)\right)
    \end{split}
\end{equation}
$\xi$ denotes the flat operation and $\oplus$ means concatenation. The input of MLP is the combination of camera rotation matrix $R_s\in \mathbb{R}^{3\times3}$, translation $t_{s}$ and camera intrinsics $K_{s}$. 


\subsection{Point-Sampling Voxel Fusion}
\label{section:voxel_fusion}
The augmented vehicle feature $f^{\prime}_{veh}$ and infrastructure feature $f^{\prime}_{inf}$ are projected into 3D space for fusion and generate two voxel features, denoted as $V_{veh}$ and $V_{inf}$, respectively. As seen in Figure~\ref{fig:Coord}, we follow the vehicle coordinate system in DAIR-V2X-C dataset to set $x,y,z$-axis of voxel volume. Each voxel has the same spatial limits in all three axes, which can be denoted as $\left[ x_{\text{min}},y_{\text{min}},z_{\text{min}},x_{\text{max}},x_{\text{max}},x_{\text{max}}\right]$ and every voxel element has the same size $\delta=(\delta_x,\delta_y,\delta_z)$. Therefore, the number of voxels along each axis can be formulated as:
\begin{equation}
N_p = \frac{p_{\text{max}}-p_{\text{min}}}{\delta_p}, p=x,y,z 
\end{equation}

We calculate the 2D coordinates $(u,v)$ in feature map $f^{\prime}_{veh/inf}$ from 3D coordinates $(x,y,z)$ in voxel volume $V_{veh/inf}$ with camera parameters. The depth $d$ at coordinates $(u,v)$ can also be calculated through the transformation.

\begin{equation}
d\left[\begin{array}{l}
u \\
v\\
1
\end{array}\right]=K_s \left[\begin{array}{ll}
R_s & t_s \\
\overrightarrow{0} & 1
\end{array}\right] \left[\begin{array}{c}
x \\
y \\
z \\
1
\end{array}\right] ,s=veh,inf
\end{equation}

All voxel elements along a camera ray are filled with the same feature following the projection principle. A voxel mask $M_s$ with the same shape as $V_s$ is defined  to indicate whether 2D coordinates are within the range of the feature map. So the Point-Sampling can be formulated as:

\begin{equation}
M_s(x, y, z)= \begin{cases}1, &  0 \leq u \leq w_0 \text{ and } 0 \leq v \leq h_0 \\ 
0, & \text { ow. }\end{cases}
\end{equation}

\begin{equation}
V_s(x, y, z)= \begin{cases}f^{\prime}_{s}(u, v), & M_s(x, y, z)=1 \\ 0, & \text { ow. }\end{cases}
\end{equation}

We obtain the final voxel feature $V_{vic} \in N_x\times N_y\times N_z \times C_1$ by averaging sampled voxel features $V_{veh}$ and $V_{inf}$. Then,  the same 3D neck as~\cite{rukhovich2022imvoxelnet}, which is composed of 3D CNN and downsampling layers, transforms voxel feature $V_{vic}$ into BEV feature $B_{vic} \in N_X \times N_y \times C_2$. BEV feature can be used as input of common 2D detection heads to predict 3D detection results.

The loss of detection heads is similar to SECOND~\cite{yan2018second}, which consists of smooth L1 Loss for bounding box $L_{\text{bbox}}$, focal loss for classification $L_{\text{cls}}$, and cross-entropy loss for direction $L_{\text{dir}}$. The final loss function can be formulated as:

\begin{equation}
L=\frac{1}{n}\left(\lambda_{\text {bbox}} L_{\text {bbox}}+\lambda_{\text {cls}} L_{\text {cls}}+\lambda_{\text {dir}} L_{\text {dir}}\right)
\end{equation}
where $n$ is the number of positive anchors.
\section{Experiments}
\subsection{Settings}

\begin{figure*}[htbp]
	\centering  
	\includegraphics[width=\linewidth]{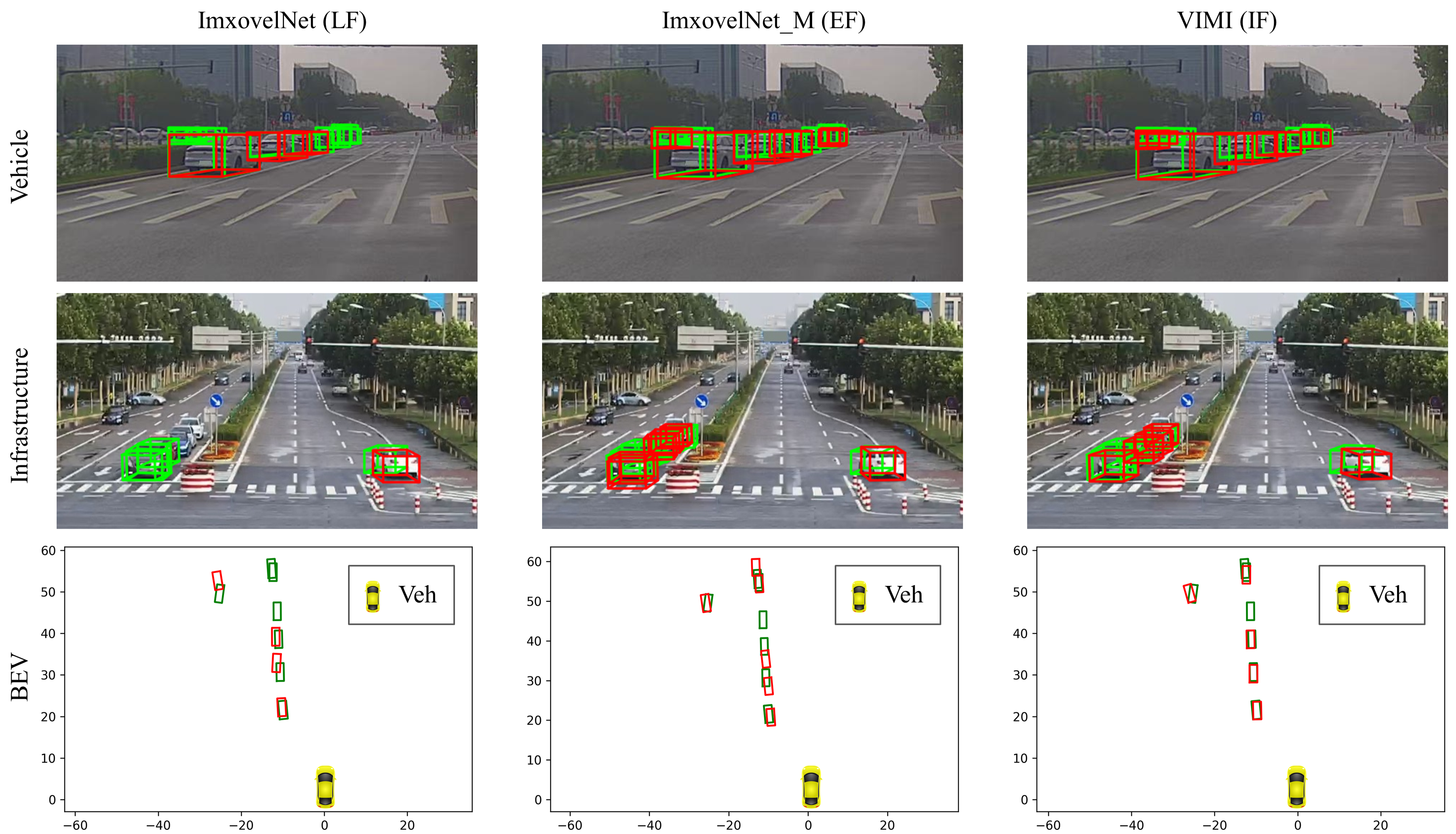} 
	\caption{Visualization results of ImVoxelNet (LF) (left column), ImVoxelNet\_M (EF) (middle column), and VIMI (IF) (right column). Bounding boxes in BEV (bottom row) are projected to vehicle and infrastructure image planes (top two rows). Groundtruth are in green and predictions in red. 
 From BEV, it is clear that red and green bounding boxes from VIMI are better aligned than LF and EF methods. This shows that ImVoxelNet (LF) and  ImVoxelNet\_M (EF) have detected more false positive objects and fewer true positive objects than VIMI (IF).
 }
	\label{fig:vis_results}   
\end{figure*}

\textbf{Datasets.}
Many studies on Vehicle-to-Infrastructure (V2I)  and Vehicle-to-Vehicle (V2V)  are based on simulated datasets, in which V2X communication systems are relatively idealistic compared with real scenarios. We conduct our experiments on multi-view 3D object detection over a vehicle-infrastructure-cooperation dataset called DAIR-V2X~\cite{yu2022dairv2x}, in which all frames are captured from real scenarios. The component dataset DAIR-V2X-C contains 38,845 frames of images and point clouds for cooperative detection tasks. We utilize the VIC-Sync portion of DAIR-V2X-C dataset for training and evaluation, which is composed of 9,311 pairs of infrastructure and vehicle frames captured at the same time. Annotations of each pair frame are in world coordinate and need to be translated into vehicle coordinate system for training and evaluation (detailed in appendix). We use the translated labels as the benchmark for our experiments. The resolution of images collected by RGB cameras is $1920\times1080$. DAIR-V2X-C dataset is split into train/val/test sets of 4822/1795/2694 frames. 


\textbf{Evaluation Metrics.}
DAIR-V2X~\cite{yu2022dairv2x} proposed to use the evaluation metrics of Average Precision (AP)~\cite{Geiger2013KITTI} and Average Byte (AB) to measure detection performance and transmission cost. AP is used to evaluate 3D detection performance compared with DAIR-V2X-C labels. Since the evaluation of VIC3D only focuses on vehicle-egocentric surroundings, we first need to filter the objects outside the cubic area $ x\in \left[x_{\text{min}},x_{\text{max}}\right]\text{m}, y \in \left[y_{\text{min}},y_{\text{max}}\right]\text{m}, z\in\left[z_{\text{min}},z_{\text{max}}\right]\text{m} $ in vehicle coordinate system. The evaluation metrics are based on the detection range surrounding the vehicle, including Overall (0-100m), 0-30m, 30-50m, and 50-100m. All metrics are calculated with $\text{IoU}=0.5$ and can be divided into 2 parts:  $AP_{\text{3D}}$ and $AP_{\text{BEV}}$. AB means the average size of transmitted data, which is the feature map $f_{inf}^{\text{T}}$ in our method. For EF and LF baselines, images and detection results are transmitted respectively.

\textbf{Baselines.} For evaluation, we compare VIMI with several baselines: $1)$ For Late Fusion, we choose ImVoxelNet~\cite{rukhovich2022imvoxelnet} as the monocular detector on each side, same as ~\cite{yu2022dairv2x}. Another LF baseline is a modified version of VIMI, named \textit{VIMI\_Veh} and \textit{VIMI\_Inf} (detailed in~\ref{detection_results}). $2)$ For Early Fusion, we consider typical methods for multi-view camera fusion in nuScenes~\cite{nuscenes2019}, such as the projection-based method BEVFormer~\cite{li2022bevformer} and the lift-based BEVDepth~\cite{li2022bevdepth}. For fair comparison, we implement BEVFormer\_S without temporal attention module. Also for compatibility with DAIR-V2X-C dataset format, we remove depth supervision from point clouds in BEVDepth to take into account camera images only. We also compare with the multi-view fusion version of ImVoxelNet~\cite{rukhovich2022imvoxelnet} (noted as \textit{ImVoxelNet\_M}). All EF methods concatenate images together and send them to backbones with shared weights.


\subsection{Implementation Details}

\textbf{Voxel Feature Construction.}  We use ResNet-50~\cite{he2016resnet} as backbone and Feature Pyramid Networks (FPN) as 2D neck to extract image features. The channel number $C$ of the neck's output is 64. We set the channel of 3D voxel feature $C_1$ to 64 and the channel of BEV feature $C_2$ to 256 following~\cite{yan2018second,lang2019pointpillars}.

We determine the spatial range of the voxel feature by considering the perception range of the camera in DAIR-V2X-C dataset. We use the same size as the Late Fusion baseline used in DAIR-V2X-C. Concretely, $\left[ x_{\text{min}},y_{\text{min}},z_{\text{min}},x_{\text{max}},y_{\text{max}},z_{\text{max}}\right]$ are set to $[0, -39.68, -3, 92.16, 39.68, 1]$ m. We set voxel resolution as $\left[0.32\times0.32\times0.33\right]$ m, so the shape of the voxel volume is $\left[288\times248\times12\right]$.

\textbf{Training.} The original image size is $\left[1080 \times 1920 \right]$ for both infrastructure and vehicle on DAIR-V2X-C dataset. In the training process, we adopt data augmentations including random scaling with the range of $\left[\text{H}\in(912,1008),\text{W}\in(513,567)\right]$ and random horizontal flipping with the probability of 50\% following~\cite{rukhovich2022imvoxelnet}.

We use AdamW optimizer with an initial learning rate of 0.0001 and weight decay of 0.0001. We use Nvidia Tesla A30 GPUs for training with batch size set to 4. The implementation is based on MMDetection3D framework~\cite{mmdet3d2020,mmdetection}, with its default training settings. VIMI is trained for 12 epochs, and learning rate is reduced by ten times after the 8-th and 11-th epoch. we use RepeatDataset in MMDetection3D involving each scene three times in one training epoch. The outputs of anchor-based detection head are filtered with Rotated NMS~\cite{mmdet3d2020} to get the final prediction.
For hyper-parameters, $\lambda_{\text {bbox}}=2.0$, $\lambda_{\text {cls}}=1.0$, $\lambda_{\text {dir}}=0.2$.

For ablation study, all experiments are trained for 12 epochs with batch size 2. As baselines of early fusion, BEVDepth~\cite{li2022bevdepth} and BEVformer\_S~\cite{li2022bevformer} are trained on DAIR-V2X-C for 20 epochs without CBGS strategy~\cite{zhu2019CGBS}.

\subsection{Object Detection Results}
\label{detection_results}
We compare the performance of baseline Late Fusion (LF) methods with ImVoxelNet and our proposed single-side model \textit{VIMI\_Veh/Inf} on DAIR-V2X-C dataset. We also implement several multi-view camera-based methods that have been applied to nuScenes dataset~\cite{nuscenes2019} in VIC3D scenario. The evaluation results on VIC-Sync portion of DAIR-V2X-C dataset are shown in Table~\ref{tab:vimi_SOTA} and Figure~\ref{fig:vis_results}. From the table, Intermediate Fusion (IF) method VIMI has achieved state-of-the-art performance on the multi-view camera fusion benchmark, compared with other methods of Late Fusion (LF) and Early Fusion (EF). VIMI obtains 15.61 $AP_{\text{3D}}$ and 21.44 $AP_{\text{BEV}}$ in overall setting.

\textit{VIMI\_Veh} and \textit{VIMI\_Inf} remove the MCA module but preserve CCM and FC modules so that models can be applied to the vehicle side and infrastructure side respectively without interaction between them, and predictions can be used for Late Fusion. VIMI achieves higher $AP_{\text{3D}}$ and $AP_{\text{BEV}}$ compared with ImVoxelNet~\cite{rukhovich2022imvoxelnet} under the setting of Only-Veh, Only-Inf, and LF. This indicates that VIMI's  single-side model has a stronger feature extraction ability.


What is interesting is that Only-Inf methods achieve the best scores in 50-100m $AP_{\text{3D}}$ and $AP_{\text{BEV}}$. As mentioned before, these metrics are related to detecting objects far from the ego vehicle. We count 16,934 objects within the distance range of 50-100m from vehicle, which are used to calculate  the metric of 50-100m $AP_{\text{3D}}$. Among these objects, almost three quarters (12,651) objects are closer to infrastructure camera, which are easier to be detected by Only-Inf models. 


\begin{table*}
  \footnotesize
  \centering
    \begin{tabular}{ccccccccccc}
    \hline
    \multirow{2}{*}{\textbf{Fusion}} & \multirow{2}{*}{\textbf{Model}} & \multicolumn{4}{c}{\bm{$AP_{\textbf{3D (IoU=0.5)}}$}} & \multicolumn{4}{c}{\bm{$AP_{\textbf{BEV (IoU=0.5)}}$}} & 
    \multirow{2}{*}{\makecell[c]{AB\\(Byte)}} \\ \cline{3-10}
                                &               & Overall   & 0-30m    & 30-50m    & 50-100m   & Overall   & 0-30m     & 30-50m    & 50-100m   & \\ \hline
    \multirow{2}{*}{Only-Veh}   & ImvoxelNet~\cite{rukhovich2022imvoxelnet}    & 7.29      & 16.98     & 2.35      & 0.13      & 8.85      & 19.89     & 3.44      & 0.28      & 
                                \multirow{2}{*}{\textbackslash{}}       \\
                                & VIMI\_Veh  & 8.65     & 19.11     & 4.33      & 0.20      & 10.46     & 22.42     & 5.57      & 0.42      & \\ \hline
    \multirow{2}{*}{Only-Inf}   & ImvoxelNet~\cite{rukhovich2022imvoxelnet}    & 8.66      & 13.05     & 5.79      & \underline{5.50}      & 14.41     & 17.98     & 10.34     & \underline{11.19}     &    
                                \multirow{2}{*}{\textbackslash{}}        \\ 
                                & VIMI\_Inf  & 9.76     & 13.59     & 6.90      & \textbf{6.63} & 14.81 & 18.78     & \underline{11.50}     & \textbf{11.43} &  \\ \hline
    \multirow{2}{*}{Late-Fusion (LF)} & ImVoxelNet~\cite{yu2022dairv2x} & 11.08  & 22.27     & 4.40      & 2.33      & 14.76     & 27.02     & 7.13      & 4.73      & 0.28K \\
                                & VIMI\_Veh/Inf & 11.99 & \underline{24.79}     & 6.08      & 2.30      & 15.79     & 30.39     & 8.50      & 4.84      & 0.28K \\ \hline
    \multirow{3}{*}{Early-Fusion (EF)} & BEVDepth~\cite{li2022bevdepth} & 7.36    & 16.23     & 1.79      & 0.18      & 13.17     & 26.42     & 5.00      & 4.82      &
                                \multirow{3}{*}{550.84K}            \\
                                & BEVFormer\_S~\cite{li2022bevformer}     & 8.80      & 18.07     & 3.71      & 1.76      & 13.45     & 24.76     & 6.46      & 4.63      &       \\
                                & ImVoxelNet\_M~\cite{rukhovich2022imvoxelnet}    & \underline{12.72}     & 23.63     & \underline{7.38}      & 3.11      & \underline{18.17}     & \underline{30.54}     & 11.39     & 7.00      &        \\ \hline
    Intermediate-Fusion (IF)    & VIMI   & \textbf{15.61} & \textbf{29.12} & \textbf{9.07} & 4.01    & \textbf{21.44} & \textbf{36.24}   & \textbf{13.51}     & 8.28      & 32.64K  \\ \hline
    \end{tabular}
    \caption{Quantitative evaluation on DAIR-V2X-C. Best values are marked by bold, and the second best is underlined. All scores in $\%$. 
    }
    \label{tab:vimi_SOTA}
\end{table*}

\subsection{Ablation Study}

We remove MCA, CCM, and FC modules in VIMI and regard it as baseline in the ablation study. We also conduct experiments to investigate when to fuse information from vehicle and infrastructure (details of compared architectures are provided in appendix).

\begin{table}[htbp]
  \footnotesize
  \centering
    \begin{tabular}{ccccc}
    \hline
    \textbf{MCA} & \textbf{CCM} & \textbf{FC} & \bm{$AP_{\textbf{3D}}$} & \bm{$AP_{\textbf{BEV}}$} \\ \hline 
            &               &               & 13.60 & 20.05   \\ 
                & \Checkmark    &               & 13.98 & 20.23   \\ 
     \Checkmark  &               &               & 14.65 & 20.64   \\
     \Checkmark  &  \Checkmark    &               & \underline{15.27} & \underline{21.03}  \\
     \Checkmark  &  \Checkmark    &  \Checkmark    & \textbf{15.61} & \textbf{21.44}   \\ \hline
     
    \end{tabular}
    \caption{Ablation study on each component of VIMI.  
    }
    \label{TAB:AB}
\end{table}

\textbf{Effect of Each Component.} The ablation results on MCA, CCM, and FC modules are summarized in Table~\ref{TAB:AB}. Comparing the 2nd and 3rd rows with the 1st row, both MCA and CCM can improve performance over baseline, and MCA has increased $AP_{\text{3D}}$ and $AP_{\text{BEV}}$ by 1.05 and 0.59, better than 0.38 and 0.18 increase induced by CCM module. These results demonstrate the validity of MCA, which selects more useful infrastructure features at different scales based on vehicle features with a cross-attention mechanism. FC is designed to eliminate redundant information included in features, while it also improves detection performance. This is because FC module increases the depth of the whole network and introduces extra computation, which can be regarded as feature-refinement.

\textbf{Early or Intermediate Fusion?} The main difference between EF and IF is the type of information transmitted. EF transmits images while IF transmits features. For the former, features can be extracted from the shared-weights backbone, which is a common method in multi-view works~\cite{li2022bevdepth,li2022bevformer,huang2022bevdet4d}. For the latter, since the views of vehicle and infrastructure cameras are different, it is reasonable that images are processed by the backbone and neck on each side separately. The output feature can be transmitted and then fused, which is a common pipeline of IF method. Meanwhile, vehicles and infrastructure can change different backbones for training. Table~\ref{TAB:VoxelBEV} shows that VIMI fusing feature at voxel level achieves better performance compared with EF.

\textbf{Voxel or BEV Fusion?} To investigate when to fuse features in IF method (at voxel or BEV level), we compare the performance of VIMI with \textit{VIMI\_BEV}.
The former belongs to the IF-Voxel pipeline while the latter belongs to the IF-BEV fusion pipeline, which condenses voxel features $V_{veh}$ and $V_{inf}$ into BEV feature respectively with two 3D necks, and then two BEV features are averaged for fusion. Results (Table~\ref{TAB:VoxelBEV}) show that fusion at the voxel level has better performance, which indicates that the transformation from voxel to BEV feature can cause higher information loss.

\begin{table}[htb]
  \footnotesize
  \centering
    \begin{tabular}{cccc}
    \hline
    \textbf{Fusion} & \textbf{Model} & \bm{$AP_{\textbf{3D}}$} & \bm{$AP_{\textbf{BEV}}$}  \\ \hline
            LF       & ImVoxelNet & 11.08  & 14.76   \\ 
            EF       & ImVoxelNet\_M & \underline{12.72}  & \underline{18.17}   \\ 
            IF (BEV)  & VIMI\_BEV & 11.50  & 16.23   \\ 
            IF (Voxel) & VIMI & \textbf{13.37}  & \textbf{19.66}   \\ \hline
    \end{tabular}
    \caption{Analysis on the choice of feature fusion. 
    }
    \label{TAB:VoxelBEV}
\end{table}

\subsection{Impact of Feature Compression}

\begin{figure}[ht]
	\centering  
	\includegraphics[width=0.9\linewidth]{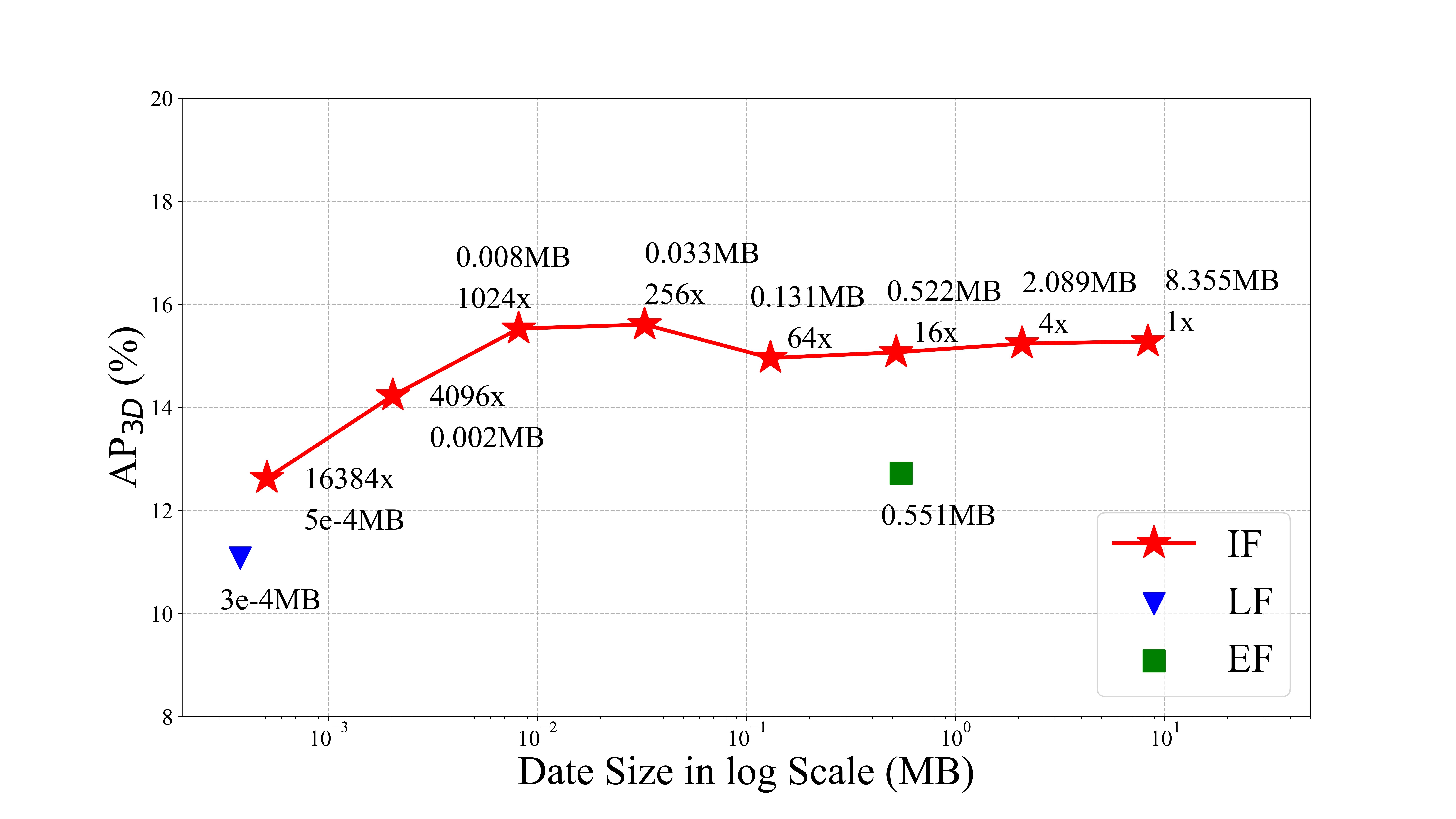} 
	\caption{$AP_{\text{3D (IoU=0.5)}}$ with respect to Compression Rate (shown as number $\times$). CCR is changed from $\times 1$ to $\times64$ and SCR is set from $\times 1$ to $\times 256$ with CCR set to $\times64$. 
 }  
	\label{fig:comp_3d}   
\end{figure}

As seen in Figure~\ref{fig:comp_3d}, We investigate the effect of Channel Compressor and Spatial Compressor. First, we change Channel Compression Rate (CCR) from $\times1$ to $\times64$, and the model performance is almost stable at low compression rates, which indicates that channel compression can extract more useful information and remove redundancy. 



After CCR reaches the maximum, we continue to compress features with Spatial Compressor. The compression rate ranges from $\times64$ to $\times 16384$. With compressed feature shapes getting smaller, the $AP_{\text{3D}}$ declines from 15.33 to 12.63 but is still higher than LF, and the transmission cost has fallen to 0.51KB which is comparable to LF's cost. 


\subsection{Analysis of Translation Noise}

In order to evaluate model robustness, we conduct several experiments to test the model's ability to overcome transmission noise.  Object position deviation exists commonly in V2X systems due to calibration noise, time asynchrony of sensor triggering, and transmission latency. Since objects like vehicles only move on the ground plane in most scenarios, this deviation can be simplified and formulated as additional noise on $x,y$ axis according to the groundtruth position. Stochastic noises $(\Delta t_x,\Delta t_y)$ simulated through Gaussian distribution with different noise amplitudes $T$ are respectively added to the part of translation $(t_x,t_y)$ of camera parameters. The standard deviation of Gaussian distribution is set to $\frac{1}{3}$ according to 3-sigma rule.

\begin{table}[htbp]
  \footnotesize
  \centering
      \resizebox{\linewidth}{!}{
    \begin{tabular}{ccccccc}
    \hline
    \multirow{2}{*}{\textbf{Metric}} & \multirow{2}{*}{\textbf{Model}} & \multicolumn{5}{c}{\textbf{Translation Noise Amplitude $T$ (m)}}
     \\ \cline{3-7} & & 0.0 & 0.1 & 0.2 & 0.5 & 1.0  \\ \hline
                            \multirow{3}{*}{$AP_{\text{3D}}$}& ImVoxelnet (LF) & 11.07 & 11.06 &11.00 & 10.76 & 9.92  \\ 
                            & VIMI\_B& 13.60 & 13.62 & 13.71 & 12.90 & 11.04   \\ 
                           & VIMI & \textbf{15.61} & \textbf{15.53} & \textbf{15.38} & \textbf{14.51} & \textbf{12.36}  \\ \hline

                          \multirow{3}{*}{$AP_{\text{BEV}}$} & ImVoxelnet (LF) & 14.89 & 14.85 & 14.84 & 14.41 & 13.32  \\ 
                            & VIMI\_B&20.05 & 20.04 & 19.90 & 18.68 & 15.87  \\ 
                           & VIMI & \textbf{21.44} & \textbf{21.43} & \textbf{21.33} & \textbf{20.46} & \textbf{17.46}  \\ \hline
    \end{tabular}}
    \caption{Results of $AP_{\text{3D (IoU=0.5)}}$ and $AP_{\text{BEV (IoU=0.5)}}$ considering transmission noise.}
    \label{TAB:T_NOISE}
\end{table}

We compare the impact of transmission noise on three models: ImVoxelNet (LF), VIMI, and \textit{VIMI\_B}, which removes MCA, FC, and CCM modules and only keeps the fusion methodology at feature level. As shown in Table~\ref{TAB:T_NOISE}, the performance of three models have a similar decline trend with noise amplitude increasing from 0.1m to 1m. However, even with noise amplitude at 1m, IF-based method VIMI achieves 12.36 $AP_{\text{3D}}$ and 17.46 $AP_{\text{BEV}}$, which are better compared to LF's results without noise (11.07 $AP_{\text{3D}}$ and 14.89 $AP_{\text{BEV}}$). The results also indicate that transmission noise has a negative impact on detection performance, and further study is needed to tackle this practical challenge.

\section{Conclusion}

VIMI is a novel multi-view intermediate-fusion framework for camera-based VIC3D task. To correct the negative effect of calibration noises and time asynchrony, we design a Multi-scale Cross-Attention module and Camera-aware Channel Masking module to fuse and augment multi-view features. VIMI also effectively reduces transmission cost via Feature Compression, and has achieved state-of-the-art results on DAIR-V2X-C benchmark, significantly outperforming previous EF and LF methods. Future study points to extension of the framework to more data modalities.

\clearpage

{\small
\bibliographystyle{ieee_fullname}
\bibliography{egbib}
}

\end{document}